\documentclass[10pt,twocolumn,letterpaper]{article}

\usepackage{cvpr}              

\definecolor{cvprblue}{rgb}{0.21,0.49,0.74}
\usepackage[pagebackref,breaklinks,colorlinks,allcolors=cvprblue]{hyperref}

\usepackage[utf8]{inputenc} 
\usepackage[T1]{fontenc}    
\usepackage{hyperref}       
\usepackage{url}            
\usepackage{booktabs}       
\usepackage{amsfonts}       
\usepackage{nicefrac}       
\usepackage{microtype}      
\usepackage{xcolor}         
\usepackage{graphicx}
\usepackage{array} 
\usepackage{colortbl}
\usepackage{bookmark}
\usepackage{wrapfig}
\usepackage{amsmath}
\usepackage{float}
\newcommand{\my}{DiC}

\def\paperID{2648} 
\def\confName{CVPR}
\def\confYear{2025}

\title{DiC: Rethinking Conv3x3 Designs in Diffusion Models}

\author{
  Yuchuan Tian$^{1*}$, Jing Han$^{2}\thanks{Equal Contribution.}$  , Chengcheng Wang$^{3}$, Yuchen Liang$^{4}$, Chao Xu$^1$, Hanting Chen$^{3\dagger}$\\
  \small$^1$ State Key Lab of General AI, School of Intelligence Science and Technology, Peking University.\\ 
  \small$^2$ School of Artificial Intelligence, Beijing University of Posts and Telecommunications. $^3$ Huawei Noah's Ark Lab. \\
  \small$^4$ School of Mathematical Sciences, Peking University.\\
  \small\texttt{tianyc@stu.pku.edu.cn, hj19910619@163.com, chenhanting@huawei.com} \\ 
}

\begin{document}
\maketitle

\begin{abstract}
  Diffusion models have shown exceptional performance in visual generation tasks. Recently, these models have shifted from traditional U-Shaped CNN-Attention hybrid structures to fully transformer-based isotropic architectures. While these transformers exhibit strong scalability and performance, their reliance on complicated self-attention operation results in slow inference speeds. Contrary to these works, we rethink one of the simplest yet fastest module in deep learning, 3x3 Convolution, to construct a scaled-up purely convolutional diffusion model. We first discover that an Encoder-Decoder Hourglass design outperforms scalable isotropic architectures for Conv3x3, but still under-performing our expectation. Further improving the architecture, we introduce sparse skip connections to reduce redundancy and improve scalability. Based on the architecture, we introduce conditioning improvements including stage-specific embeddings, mid-block condition injection, and conditional gating. These improvements lead to our proposed Diffusion CNN (\textbf{DiC}), which serves as a swift yet competitive diffusion architecture baseline. Experiments on various scales and settings show that DiC surpasses existing diffusion transformers by considerable margins in terms of performance while keeping a good speed advantage. Project page: \url{https://github.com/YuchuanTian/DiC}
\end{abstract}

\section{Introduction}

Diffusion-based image generative models have recently gained significant attention, with notable works like Stable Diffusion~\cite{sd} demonstrating impressive results. As these models continue to evolve, they are now capable of generating increasingly realistic images. Recent advancements, including Sora-related works, have extended this capability to video generation, producing high-quality, temporally consistent videos that rival real-world footage.

The self-attention mechanism plays a pivotal role in many recent diffusion models. Early work~\cite{ddpm,songunet} fused convolutional U-Net structures with self-attention. More recently, architectures such as U-ViT~\cite{uvit} and DiT~\cite{dit} have fully transitioned to attention-based designs, abandoning convolutional U-Nets entirely. These models demonstrate remarkable generative capabilities, particularly when scaled up. Newer text-to-image and text-to-video models, including Stable Diffusion 3~\cite{sd3}, FLUX~\cite{flux}, PixArt series~\cite{pixartalpha,pixartdelta,pixartsigma}, and OpenSora~\cite{opensora,opensoraplan}, have also adopted fully transformer-based architectures, achieving impressive generation quality.

\begin{figure}[!t]
  \centering
  \includegraphics[width=0.45\textwidth]{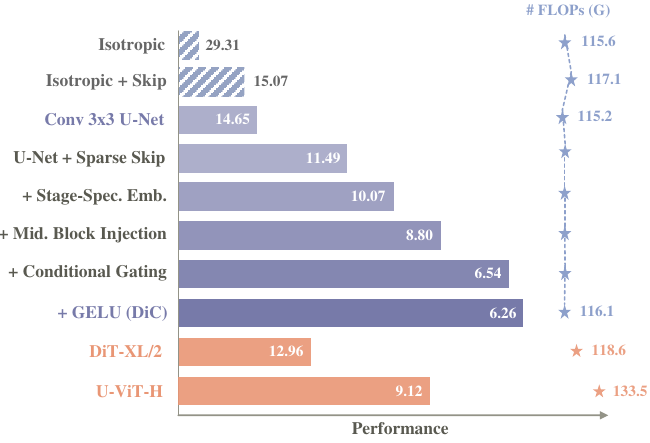}
  \vspace{-5pt}
  \caption{\textbf{The Roadmap to a 3x3 Convolutional Diffusion Model.} Performance is measured by FID ($\downarrow$) of the model trained at 200K iterations. A series of proposed model improvements gives DiC an advantage over Diffusion Transformers.}
  \label{fig:roadmap}
  \vspace{-10pt}
\end{figure}

Unfortunately, diffusion models incorporating self-attention face significant computational overhead and latency when scaled up. Complex transformer architectures pose a considerable challenge for real-time, resource-constrained applications, where large-scale transformers are impractical due to their substantial time costs.

Efforts have been made to accelerate self-attention in diffusion models. For instance, ToDo~\cite{todo} and PixArt-$\Sigma$~\cite{pixartsigma} implement more efficient self-attention mechanisms. However, these approaches remain confined to the self-attention paradigm. Other works~\cite{diffussm,dim} aim to replace self-attention with State-Space-Model-based architectures~\cite{mamba}, attempting to bypass its inherent inefficiencies. Despite these innovations, the generation speed on typical latent-domain image resolutions remains suboptimal, failing to meet the demands of real-time or large-scale applications.

In this work, we rethink the simple yet efficient fully convolutional diffusion model architecture that offers significant speed advantages. As a counterpart to self-attention, convolution is widely regarded as hardware-friendly, with excellent support across various platforms. Among the many convolutional operations, the canonical 3x3 stride-1 convolution stands out for its exceptional speed, largely due to widespread hardware optimization techniques such as Winograd acceleration~\cite{winograd}. For instance, work like RepVGG~\cite{repvgg} have demonstrated how simple convolutions can achieve both high efficiency and competitive performance.

However, 3x3 convolutions come with inherent limitations, particularly a constrained receptive field, which may hinder their scalability in complex generative tasks. We firstly perform initial trials to scale up purely 3x3-based CNNs within existing scalable frameworks, but these experiments yielded results that are still inferior to Diffusion Transformer counterparts.

To address this, we propose a series of adaptations to conventional ConvNets, forming a roadmap (shown in Fig.~\ref{fig:roadmap}) that tailors them specifically for a scaled-up diffusion model. Firstly, we focus on refining the model architecture to better leverage Conv3x3. Existing designs fall into three categories: isotropic architectures like DiT~\cite{dit}, isotropic architectures with skips like U-ViT~\cite{uvit}, and encoder-decoder hourglass structures used in canonical CNN \& self-attention hybrid models. Empirical observations reveal that the hourglass architecture is more effective for pure Conv3x3 models, because downsampling and upsampling in encoders expand the receptive field that makes up for the narrow scope of Conv3x3. Skip connections are also observed crucial, but traditional blockwise skips for cheap convolution block scale-ups are too dense to be efficient for large ConvNets. To address this, we introduce sparse skip connections that reduces the number of skips while ensuring essential information flows efficiently from encoder to decoder.

Besides, we focus on improving conditioning to fit the hourglass ConvNet better. In existing models, a single set of condition embeddings is often mapped to different blocks across various stages. However, in encoder-decoder hourglass ConvNets, each stage operates in distinct feature spaces. Hence, we introduce stage-specific embeddings, ensuring each stage uses independent, non-overlapping condition embedding tables. We also carefully inspect the position for embedding injection, and we suggest inject conditions mid-block for better performance. Additionally, we borrow conditional gating on feature-maps. In addition, we refined the internal block structure, replacing all activation functions with GELU.

With these enhancements, we propose \textbf{DiC}, a diffusion model made up of Conv3x3. The architectural and conditioning improvements collectively enable our model to achieve outstanding results while maintaining its speed advantage. Experiments on various scales and settings demonstrate DiC's performance advantage over Diffusion Transformers as well as its high-throughput characteristics.

\begin{figure*}[!t]
  \centering
  \includegraphics[width=0.9\textwidth]{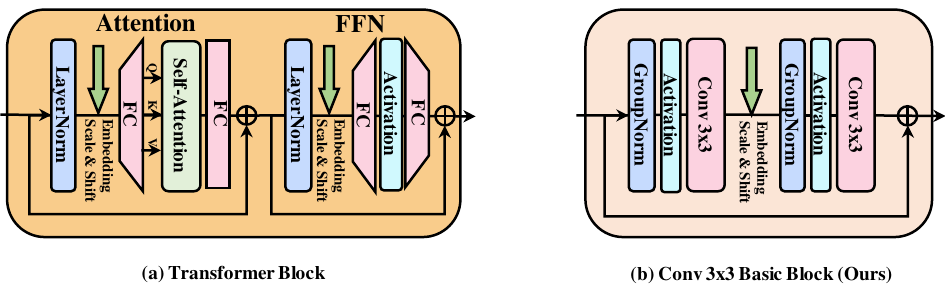}
  \vspace{-10pt}
  \caption{\textbf{Basic Blocks for scaled-up architectures.} Compared to complicated Transformer blocks widely applied in~\cite{dit}, our Conv 3x3 Basic Block is much simpler and better supported by hardware, being the key to high throughput.}
  \label{fig:block}
\end{figure*}

\section{Related Work~\label{sec:related}}
  \textbf{Self-Attention in Diffusion Architectures.} Self-attention has become a cornerstone of modern diffusion architectures, thanks to its remarkable effectiveness in capturing long-range dependencies. Early UNet-based diffusion models~\cite{ddpm,songunet} incorporated Self-Attention layers in the higher stages of the network, where capturing global context is most critical. Dhariwal et al. took this further with ADM~\cite{dhariwal}, extending self-attention across all stages, which significantly increased the model's generative capabilities.

  More recent works~\cite{uvit,dit,rin} have fully embraced transformer-based architectures, replacing traditional U-Net backbones with full self-attention designs. These models not only achieve superior performance, but also exhibit excellent scalability, particularly at large model sizes and FLOPs. Most following work like PixArt-$\alpha$~\cite{pixartalpha} and Stable Diffusion 3 adopt the full transformer backbone design that demonstrates outstanding Text-to-Image generation capabilities. Beyond these, some works continue to explore and refine the use of self-attention and transformer architectures in diffusion models, further pushing the boundaries of efficiency in transformer-based generative modeling. For instance,~\cite{todo,pixartsigma,udit} looks at downsampled tokens for self-attention;~\cite{sana} uses linear attention. Other works~\cite{diffussm,dim} opt for State Space Models for diffusion, but SSMs suffer in terms of latency for ordinary-shaped images~\cite{dim}. Some other work~\cite{simplediffusion,hourglass,edm2,udit} looks at a combination of U-Net and Self-Attention. In contrast to these works, we push architectural simplicity to the extreme by designing an entire diffusion model using 3x3 convolutions. With carefully crafted structural optimizations, this streamlined network not only capitalizes on the speed advantages of 3x3 convolutions but also achieves diffusion generation performance on par with transformer-based models.
  
  \noindent\textbf{Efficient Convolution.} Historically, architectures like AlexNet~\cite{alexnet}, VGG~\cite{vgg}, and ResNet~\cite{resnet} have dominated computer vision tasks. While transformers~\cite{transformer} have later become the state-of-the-art for many vision applications, their high computational cost and latency make them less practical in resource-constrained environments, where CNNs still excel. Recent advancements have revisited and improved CNNs, enabling them to rival or even surpass transformers in both performance and efficiency.

  For example, ConvNeXt~\cite{convnext} achieves transformer-level performance through macro and micro architectural changes, such as increasing kernel sizes and optimizing normalization. Deformable Convolutions (DCNs)~\cite{dcnv1,dcnv2,dcnv3} enhance the receptive field by learning spatially adaptive kernel shapes, while RepLKNet~\cite{replknet} utilizes massive 31x31 kernels to capture long-range dependencies. Similarly, Wavelet Conv~\cite{waveletconv} operates in the frequency domain using wavelet transforms to extend CNN capabilities.
  
  On the other hand, there is work that resort to improving the efficiency of simple convs. RepVGG~\cite{repvgg} focuses on the potential of simple 3x3 convolutions, employing re-parameterization tricks to enhance their effectiveness. Inspired by its speed potential, we explore the use of pure 3x3 convolutions in diffusion, combined with architectural and conditioning improvements, to achieve competitive results in diffusion-based generative models. However, we are not using tricks like re-parameterization~\cite{repvgg}. Our approach demonstrates that even with a simple design, high-quality generation can be both efficient and effective.

\section{Method}
\subsection{Preliminaries: Conv3x3 as the Ingredient\label{sec:3.1}}

We choose normal stride-1 full Conv3x3 as the major ingredient of model. Stride-1 3x3 convolutions are incredibly speedy, primarily due to extensive hardware and algorithmic optimizations in modern deep learning frameworks. The operation optimization benefits from the Winograd algorithm~\cite{winograd}, which reduces the number of multiplications required for common convolution operations by $\frac{5}{9}$. For 3x3 kernels, this method significantly accelerates the computation by leveraging efficient matrix transformations.

Additionally, unlike depth-wise convolution, 3x3 convolutions achieve a high degree of computational parallelism, allowing modern GPUs to fully utilize their processing power. This parallelism, combined with minimal memory access overhead, makes stride-1 3x3 convolutions particularly efficient. Compared to larger or more complex convolution operations, they offer an ideal trade-off between computational cost and representational power, making them a practical choice for real-time applications and large-scale models.

\subsection{Architectures for Scale-Up\label{sec:3.2}}

\noindent\textbf{Our Basic Block.} We revisit canonical diffusion models~\cite{songunet,dhariwal} and figure out that 3x3 convolutions are also widely applied. However, these models are \textbf{not} pure CNNs: the basic block unit within the model consists of two consecutive 3x3 convolutional layers, followed by a self-attention module that attends within the entire image. Before each convolution, the input is normalized using GroupNorm, and then passed to the non-linear activation function of SiLU. Notably, the intermediate features within the block maintain the same number of channels, ensuring stable feature propagation. The overall block adopts a residual structure~\cite{resnet}, where the input is directly added to the output via a shortcut connection.

The roadmap to our proposed diffusion model \my -- a 3x3 convolutional denoiser starts from the revisited convolutional basic block mentioned above. In order to satisfy the ultimate goal of the roadmap, we remove the self-attention layers after the convolutions to form a minimal unit block (Basic Block) consisting of two 3x3 convolutions. Other block designs are temporarily left intact.

\begin{figure}[!t]
  \centering
  \includegraphics[width=0.48\textwidth]{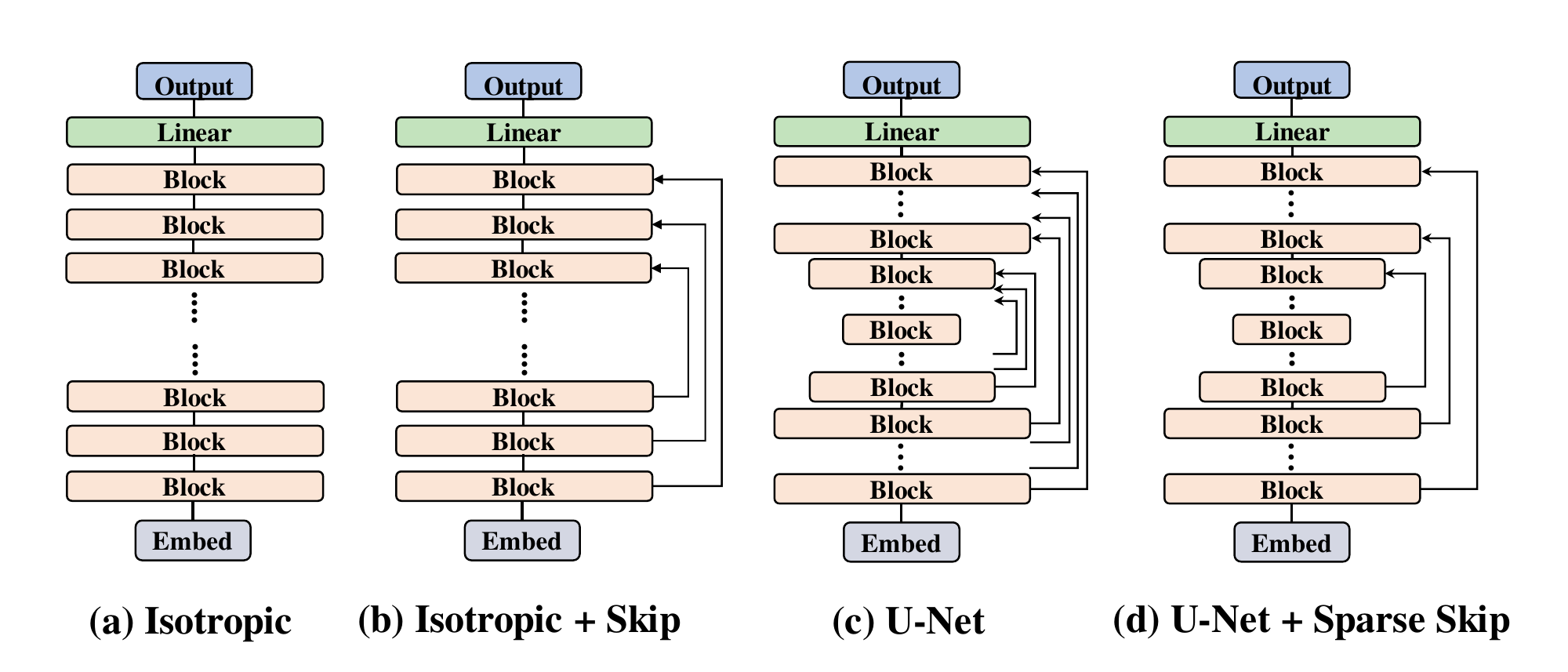}
  \vspace{-20pt}
  \caption{\textbf{Mainstream Diffusion Architectures.} We try existing diffusion architectures (a,b,c) with Conv3x3 Basic Block. We further improve U-Net (d) with sparse-skips.}
  \label{fig:arch}
  \vspace{-10pt}
\end{figure}

\noindent\textbf{Mainstream Architectures.} With this Conv 3x3 Basic Block as the foundation, we scale the model up by stacking multiple such blocks and increasing the network's depth and width, and we are faced up to numerous architectural choices (shown in Fig.~\ref{fig:arch}):

1. Isotropic Architecture. Inspired by DiT and subsequent works~\cite{dit,diffit,visionllama,pixartalpha}, this design features a simple, vertically stacked structure. After patchification, the intermediate feature-map size is not changed throughout the entire model. It is widely adopted for its reputed scalability, particularly when scaled to large models.

2. {Isotropic Architecture with Skip Connections.} This approach extends the columnar isotropic architecture by introducing long skip connections between non-adjacent layers, similar to the design used in U-ViT~\cite{uvit}. Unlike traditional U-Net, despite the skip design, this architecture maintains a consistent spatial resolution throughout the network, avoiding any upsampling and downsampling operations.

3. {U-Net Hourglass Architecture.} This classical design~\cite{unet} consists of an encoder and a decoder, forming a funnel-like shape. The encoder progressively downsamples the feature dimensions, while the decoder symmetrically upsamples the spatial dimensions. Apart from the hourglass backbone, each block in the encoder is connected to its corresponding decoder block via skip connections. This work is widely applied in earlier diffusion works~\cite{ddpm,songunet,dhariwal}.

\begin{table}[htbp]
  \centering
  \setlength{\belowcaptionskip}{0cm}   
\begin{tabular}{lccc}
  \toprule
  \multicolumn{4}{l}{\bf{ImageNet} 256$\times$256, 200K, cfg=1.5} \\
  \toprule
  Model & FLOPs (G) & FID$\downarrow$ & IS$\uparrow$ \\
  \midrule
  \textbf{DiT-XL} & 118.6 & 12.96 & 94.26 \\
  \midrule
  \textbf{Isotropic} & 115.6 & 29.31 & 52.80 \\
  \textbf{Isotropic + Skip} & 117.1 & 15.07 & 85.51 \\
  \textbf{U-Net Hourglass} & 115.2 & 14.65 & 87.17 \\
  \midrule
  \textbf{U-Net + Sparse Skip} & 116.1 & \textbf{11.49} & \textbf{106.91} \\
  \bottomrule
  \end{tabular}
  \vspace{-5pt}
  \caption{\textbf{Comparing Architectures for Conv3x3 Scale-Up on ImageNet 256$\times$256 conditional generation.} Conv3x3 taps its full potential with U-Net Hourglass architectures due to enlarged perception field; sparsifying skip connection further improves the model.}
  \label{tab:method_arch}
  \vspace{-5pt}
\end{table}

\noindent\textbf{Experimental Analysis.} Using the foundational Conv 3x3 Basic Block, we conducted scaling experiments on the aforementioned architectural choices. The results, summarized in Tab.~\ref{tab:method_arch}, indicate that the performance across all architectures was generally suboptimal. However, these experiments revealed several patterns that provide valuable insights for designing an effective 3x3 convolution-based diffusion model.

Firstly, for diffusion models composed solely of 3x3 convolutions, the encoder-decoder architecture hourglass is essential. This stems from a fundamental limitation of 3x3 convolutions: their inherently restricted receptive field. In a purely isotropic, standard-transformer architectured setup, each 3x3 convolution expands the receptive field by only one pixel in each direction. As a result, achieving a global receptive field would require a deep stack of convolutions, leading to inefficiencies in both computation and parameters. In contrast, an encoder-decoder architecture significantly mitigates this limitation. By progressively downsampling in the encoder and upsampling in the decoder, the 3x3 convolutions in higher stages can effectively perceive larger regions of the input. For instance, at deeper stages, a single 3x3 convolution can cover areas as large as 6x6 or 12x12 in the original image space, substantially increasing the model's receptive field. This hierarchical approach allows the model to efficiently capture both local and global context, making it far more effective for generative tasks.

Secondly, skip connections are vital in Conv3x3-based architectures. On one hand, in encoder-decoder structures, they mitigate information loss during downsampling by passing feature maps directly to the decoder, enriching its representation. On the other hand, skips accelerate training and improve generation quality by providing additional gradient pathways and preserving important features, leading to more efficient and effective diffusion modeling.

Above all, we adopt the U-Net as the general model architecture that we use.

\noindent\textbf{Improving U-Nets via Strided Skip.} However, as scaling up our Conv3x3-based diffusion model requires stacking numerous convolutional layers, dense, block-wise skip connections become a bottleneck. Numerous blockwise skip features have to be concatenated to the features and get merged on the channel dimension within corresponding decoder blocks, consuming excessive computational costs. This inefficiency hampers the scalability of the model. To address this, we propose strided skip connections, where skips are applied only every few blocks instead of after each one. This approach improves the performance, and our experiments (in Tab.~\ref{tab:method_arch}) show that it does yield better results compared with densely-connected skip designs.

\subsection{Conditioning Improvements\label{sec:3.3}}

Apart from architectural improvements, we hold that conditioning could also be refined. After model scale-ups, previous conditioning designs need to be improved according to the architecture.

\noindent\textbf{Stage-Specific Embeddings.} In conventional diffusion models~\cite{songunet,dhariwal}, a unique embedding table is shared by previous models. However, in the scaled-up Conv3x3 model, we rely on an encoder-decoder structure to expand the receptive field. Each stage of the encoder-decoder operates with different channel dimensions, reflecting significant structural and functional differences between stages. Given that feature dimensions vary across stages, applying the same condition embedding uniformly on all blocks from different stages may not be ideal. The diverse roles and representations at each stage suggest that a more stage-tailored approach to condition embedding could be more effective, as a single embedding set might not adequately capture the distinct characteristics of each stage.

\begin{table}[htbp]
  \centering
  \setlength{\belowcaptionskip}{0cm}   
\begin{tabular}{lccc}
  \toprule
  \multicolumn{4}{l}{\bf{ImageNet} 256$\times$256, 200K, cfg=1.5} \\
  \toprule
  Model & FLOPs (G) & FID$\downarrow$ & IS$\uparrow$ \\
  \midrule
  \textbf{Sparse-Skip U-Net} & 116.1 & 11.49 & 106.91 \\
  \textbf{+ Stage-Spec. Emb.} & 116.1 & 10.07 & 121.30 \\
  \textbf{+ Mid-Block Injection} & 116.1 & 8.80 & 134.25 \\
  \textbf{+ DiT condition gating} & 116.1 & 6.54 & 162.34 \\
  \midrule
  \textbf{DiC (+ GELU)} & 116.1 & \textbf{6.26} & \textbf{170.04} \\
  \bottomrule
  \end{tabular}
  \vspace{-5pt}
  \caption{\textbf{Conditioning improvements on ImageNet 256$\times$256 conditional generation.} The proposed conditioning improvements collectively bring considerable performance enhancement. Experiments are conducted using hyperparameters from~\cite{dit} for 200K iterations.}
  \label{tab:method_conditioning}
  \vspace{-10pt}
\end{table}

As an improvement, we propose using stage-specific condition embeddings for each stage of the encoder-decoder structure. Each stage will have its own independent embedding, with the embedding dimension aligned to that stage's feature dimension. This ensures that the condition embedding is tailored to the specific characteristics and scale of each stage.

It is worth mentioning that we analyzed the overhead introduced by the stage-specific embeddings. The results show that the added overhead is minimal. Specifically, adding the new stage-specific embeddings increased the model size by just 14.06M parameters, which accounts for only 2\% of the total model size. The computational overhead introduced is even more negligible, with an increase of only 12M FLOPs, which has little impact on the overall computational efficiency. This demonstrates that the inclusion of stage-specific embeddings does not significantly affect the model's performance or computational cost.

Beside, the label-drop operation on the embeddings across different stages should be synchronized during training. This prevents any leakage of label between stages and ensures that the model learns in a consistent and isolated manner, thereby improving the stability and performance of the diffusion model.

\noindent\textbf{Condition Injection Position.} Another critical aspect of our design is determining the optimal position for injecting conditioning information. There are two prevalent strategies for condition injection in diffusion models. The first approach injects the condition at the very beginning of each block, typically through LayerNorm, as seen in models like~\cite{dit,visionllama}. The second approach introduces the condition in the middle of the block, as utilized in~\cite{songunet,dhariwal}.

Through experimentation, we found that for our fully convolutional diffusion architecture, injecting the condition into the second convolutional layer within each block yields the best performance. This placement effectively modulates the feature representations, enhancing the model's generative quality without compromising its efficiency.

\noindent\textbf{Conditional Gating.} In addition, we adopted the conditional gating mechanism from DiT's AdaLN~\cite{dit}. On top of traditional affine conditioning that performs channel-wise affine mapping to the feature map, AdaLN introduces a gating vector that scales the features along the channel dimension, providing a more dynamic and fine-grained control over the conditioning process. This modification enhances the model's ability to adapt to different conditions and improves the overall generation quality. Our experiments show that this enhancement yields a notable performance boost, further demonstrating the effectiveness of this conditioning strategy in diffusion models.

\subsection{Minor Modification\label{sec:3.4}}

We re-visit the activation function across the entire model. Inspired by ConvNeXt~\cite{convnext}, we replaced the commonly used SiLU activation with GELU, which is a standard choice in transformer architectures. On top of other modifications, this change contributed to a modest but consistent improvement in performance. We are aware of potentially better activations~\cite{swish,enhancing}, but we stick to GeLU for simplicity.

\section{Experiments\label{sec:experiments}}
  
In this section, we provide experiments on various settings to demonstrate the outstanding performance of \my~model.

\begin{table*}[htbp]
	\centering
	\setlength{\belowcaptionskip}{0cm}
	\begin{tabular}{ccccccc}
		\toprule
		Model & Params (M) & FLOPs (G) & Wino. FLOPs (G) & Channel & \# Groups & Encoder-Decoder \\
		\midrule
		\textbf{\my-S} &  32.8 & 5.9 & 2.9 & 96 & 16 & [6,6,5,6,6]\\
		\textbf{\my-B} & 129.5 & 23.5 & 11.8 & 192 & 32 & [6,6,5,6,6] \\
		\textbf{\my-XL} & 702.3 & 116.1 & 57.2 & 384 & 32 & [7,7,8,7,7] \\
		\textbf{\my-H} & 1034.4 & 204.4 & 97.2 & 384 & 32 & [14,14,10,14,14] \\
		\bottomrule
	\end{tabular}
	\vspace{-5pt}
	\caption{\textbf{Configurations of DiC architecture with different model sizes.} We align DiC architectures to DiTs in terms of both FLOPs and parameters. Wino. FLOPs takes the FLOPs saved by Winograd into consideration. \# Groups stands for the number of groups in GroupNorm. Encoder-Decoder denotes the transformer block number of encoder and decoder module.}
	\label{tab:configurations}
  \vspace{-15pt}
\end{table*}

\subsection{Experiment Setups}

\textbf{Model Configs.} To benchmark against existing DiT~\cite{dit} models, we designed our architectures at different scales: Small (S), Big (B), and Extra Large (XL), which align closely with DiT-S/2, DiT-B/2, and DiT-XL/2 in terms of both FLOPs and parameter count, as shown in Table \ref{tab:configurations}. Additionally, we introduce a larger model, Huge (H), for further scalability testing. With 32.8M parameters and 5.9G FLOPs, \textbf{\my-S} provides a lightweight solution, similar to DiT-S/2, with an efficient [6,6,5,6,6] encoder-decoder configuration.
\textbf{\my-B} scales up to 129.5M parameters and 23.5G FLOPs, which is comparable to DiT-B/2, leveraging a wider channel size of 192. \textbf{\my-XL} has 702.3M parameters and 116.1G FLOPs, whose configuration rivals DiT-XL/2, with an enhanced encoder-decoder depth of [7,7,8,7,7]. \textbf{\my-H} is further scaled up to [14,14,10,14,14] and is designed to evaluate the potentials of DiC models.

Apart from normal FLOPs calculation, in Tab.~\ref{tab:configurations} we also record actual FLOPs when taking Winograd acceleration~\cite{winograd} into account. As is introduced in Sec.~\ref{sec:3.1}, Winograd could reduce the computation of a stride-1 3x3 Convolution by 5/9. From statistics in Tab.~\ref{tab:configurations}, it could be observed that Winograd could reduce the amount of theoretical FLOPs of the proposed DiC by approximately a half.

\noindent\textbf{Experiment Settings.} We follow the standard training setup from previous research~\cite{dit,sit,visionllama,udit}, utilizing the same VAE (\textit{i.e.}, sd-vae-ft-ema) as in latent diffusion models~\cite{sd} and the AdamW optimizer. All hyperparameters remain unchanged, including a global batch size of 256, a learning rate of $1e^{-4}$, weight decay of 0, and a global seed of 0. Training is performed on the ImageNet 2012 dataset~\cite{imagenet}. Our DiC-XL and DiC-H models are trained on 8 Ascend 910B NPUs.

\noindent\textbf{Overhead Calculation.}
We calculate the FLOPs overhead of \my~models via torchprofile~\cite{torchprofile}, an easy-to-use but comprehensive FLOPs calculator. The throughput is calculated on a single A100. Following the official sampling configurations of~\cite{dit}, we use batchsize 32 for throughput testing. We heat the GPU up via numerous warm-up runs before measuring the throughput. Flash attention~\cite{dao2022flashattention,dao2023flashattention2} is activated based on the original implementations of baselines.

\subsection{Evaluating DiC on the Standard DiT Setting}

\begin{table}[htbp]
  \centering
  \setlength{\belowcaptionskip}{0cm}   
\begin{tabular}{lccc}
  \toprule
  \multicolumn{4}{l}{\bf{ImageNet} 256$\times$256, 400K} \\
  \toprule
  Model & FLOPs (G) & FID$\downarrow$ & IS$\uparrow$ \\
  \midrule
  \textbf{DiT-S/2} & 6.1 & 67.40 & 20.44 \\
  \textbf{\my-S (Ours)} & 5.9 & \textbf{58.68} & \textbf{25.82} \\
  \midrule
  \textbf{DiT-B/2} & 23.0 & 42.84 & 33.66 \\
  \textbf{\my-B (Ours)} & 23.5 & \textbf{32.33} & \textbf{48.72} \\
  \midrule
  \textbf{DiT-XL/2} & 118.6 & 20.05 & 66.74 \\
  \textbf{\my-XL (Ours)} & 116.1 & \textbf{13.11} & \textbf{100.15} \\
  \bottomrule
  \end{tabular}
  \vspace{-5pt}
  \caption{\textbf{Comparing DiCs against DiTs across various model sizes.} We compare models trained for 400K iterations with the standard setting of DiT. DiCs at all scales outperform DiTs at considerable margins.}
  \label{dicvsdit}
  \vspace{-5pt}
\end{table}

\noindent\textbf{Comparison with DiT Across Model Sizes.} As shown in Table \ref{dicvsdit}, our DiC models outperform the corresponding DiT models across all tested scales. Specifically, \my-S achieves a significant reduction in FID, dropping from 67.40 to 58.68, while also surpassing DiT-S/2 in IS, increasing from 20.44 to 25.82. Similarly, \my-B outperforms DiT-B/2 by a notable margin, reducing FID from 42.84 to 32.33 and increasing IS from 33.66 to 48.72.

In the large-scale model category, \my-XL delivers a remarkable improvement over DiT-XL/2, with FID dropping from 20.05 to 13.11 and IS soaring from 66.74 to 100.15. These results highlight the effectiveness of our DiC models, which consistently achieve superior performance in both image quality and diversity, all while maintaining computational efficiency.

\begin{figure*}[htbp]
  \begin{minipage}{0.48\textwidth}
    \centering
    \includegraphics[width=0.9\textwidth]{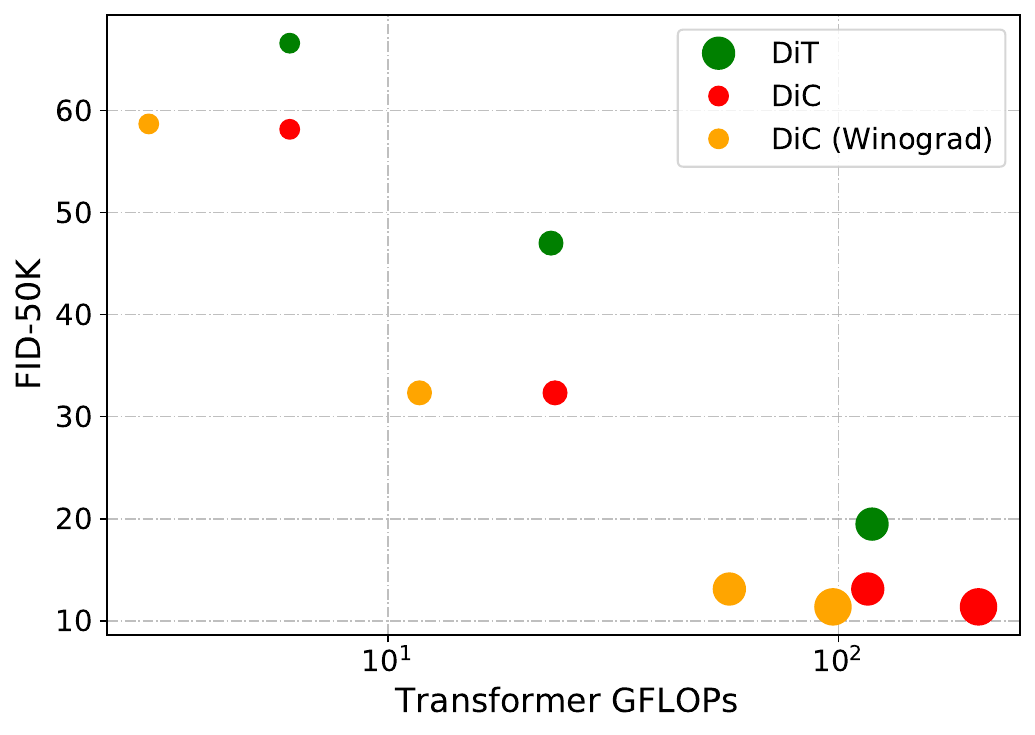}  
    \caption{\textbf{Comparing FID performance and FLOPs between DiTs and DiCs.} Marker sizes represents relative model sizes.}
    \label{fig:sample}
  \end{minipage}
  \qquad
  \begin{minipage}{0.48\textwidth}
    \centering
    \resizebox{\textwidth}{!}{
      \centering
    \begin{tabular}{lcccc}
      \toprule
      \multicolumn{5}{l}{\bf{ImageNet} 256$\times$256, 400K} \\
      \toprule
      Model & FLOPs (G) & TP & FID$\downarrow$ & IS$\uparrow$ \\
      \midrule
      \textbf{U-ViT-XL}~\cite{uvit} & 113.0 & 72.6 & 18.35 & 76.59 \\
      \textbf{DiT-XL/2}~\cite{dit} & 118.6 & 66.8 & 20.05 & 66.74 \\
      \textbf{PixArt-$\alpha$-XL/2}~\cite{pixartalpha} & 118.4 & 64.1 & 24.75 & 52.24 \\
      \textbf{DiffiT-XL/2}~\cite{diffit} & 118.5 & 64.1 & 36.86 & 35.39 \\
      \textbf{DiT-LLaMA$^*$}~\cite{visionllama} & 118.6 & 65.2 & 20.22 & 70.10 \\
      \midrule
      \textbf{\my-XL (Ours)} & 116.1 (57.2) & \textbf{313.7} & 13.11 & 100.15 \\
      \textbf{\my-H (Ours)} & 204.4 (97.2) & 160.8 & \textbf{11.36} & \textbf{106.52} \\
      \bottomrule
    \end{tabular}
    }
    \vspace{-5pt}
    \captionof{table}{\textbf{The performance of DiCs and competitive diffusion architectures on ImageNet 256$\times$256 generation.} The baselines are aligned under the official 400K-iteration setting of DiT-XL for a fair comparison. TP stands for throughput. FLOPs considering Winograd is reported for DiCs.}
    \label{dicvsother}
  \end{minipage}
\end{figure*}

\noindent\textbf{Comparison with Diffusion Transformer Baselines.} 
As different models use different settings, including training hyperparameters, the choice of samplers, training iterations et cetera, we adopt a universally-aligned setting (400K iterations on the DiT codebase) according to~\cite{udit}. As shown in Table \ref{dicvsother}, our proposed DiC models consistently outperform baseline diffusion architectures on ImageNet 256x256, all under similar computational budgets. Despite having comparable FLOPs to models like DiT-XL/2 and PixArt-$\alpha$-XL/2, our DiC-XL achieves a significantly lower FID of 13.11 and an impressive IS of 100.15, showcasing its superior generative quality.

Beyond performance, another standout is DiC's efficiency in speed. Despite the heavy computational demands of diffusion models, DiC-XL achieves a throughput of 313.7, which, while appearing lower, reflects the use of Conv3x3's highly optimized operations. This architecture capitalizes on the inherent speed of simple convolutions, offering a practical balance between efficiency and quality.

Even more impressively, DiC-H strikes a remarkable trade-off between model size and speed, delivering state-of-the-art FID of 11.36 and an IS of 106.52, with an improved throughput of 160.8. This demonstrates that our models can achieve unparalleled generation quality without the severe latency typically associated with transformer-based diffusion models. In real-world applications, these speed advantages make DiC a practical solution for high-quality image synthesis.

\begin{table}[htbp]
  \centering
  \setlength{\belowcaptionskip}{0cm}   
\begin{tabular}{lcccc}
  \toprule
  \multicolumn{5}{l}{\bf{ImageNet} 256$\times$256, 400K} \\
  \toprule
  Model & cfg & FLOPs & FID$\downarrow$ & IS$\uparrow$ \\
  \midrule
  \textbf{DiT-XL/2} & 1.5 & 118.6 & 6.24 & 150.10 \\
  \textbf{U-ViT-XL} & 1.5 & 113.0 & 5.66 & 170.62 \\
  \textbf{DiC-XL} & 1.5 & 116.1 (57.2) & \textbf{3.89} & \textbf{224.20} \\
  \bottomrule
  \end{tabular}
  \vspace{-5pt}
  \caption{\textbf{Generation performance with classifier-free guidance.} The proposed DiC is also performant on conditional generation.}
  \label{dicvsdit_cfg}
  \vspace{-15pt}
\end{table}

Furthermore, we experiment the conditional generation capabilities of DiC model in Tab.~\ref{dicvsdit_cfg}. DiC-XL achieves an FID of 3.89 at merely 400K iterations, surpassing competitive Diffusion Transformer baselines by large margins. The results are yielded from conditional generation samples with cfg=1.5.

\subsection{DiC on Larger Images}

As shown in Table \ref{tab:imagenet512}, we compare DiC models with DiT on ImageNet 512x512, after training for 400K iterations with standard DiT hyperparameters. The key takeaway from these results is the computational efficiency of DiC models. While DiT-XL/2 requires 524.7G FLOPs (with Winograd optimization), our DiC-XL model performs significantly better in terms of FID and IS, with only 464.3G FLOPs (228.7G after Winograd optimization). This improvement becomes even more evident in DiC-H, which offers a dramatic reduction in FID (12.89) and achieves a higher IS (101.78) with fewer FLOPs than DiT-XL/2.

The key difference in computational complexity between DiC and DiT lies in their underlying architectures. DiC uses a pure convolutional approach, which scales linearly with the image size. In contrast, DiT employs self-attention, which incurs quadratic complexity with respect to the image size. Therefore, for larger images, such as 512x512, the computational gap between DiC and DiT becomes more pronounced. In spite of this, we are amazed that DiC models could still maintain an advantagous gap over DiTs. The advantage of DiC, on the other hand, is also outstanding in terms of throughput.

\begin{table}[htbp]
  \centering
  \setlength{\belowcaptionskip}{0cm}   
\begin{tabular}{lcccc}
  \toprule
  \multicolumn{5}{l}{\bf{ImageNet} 512$\times$512, 400K} \\
  \toprule
  Model & G FLOPs (Wino.) & TP & FID$\downarrow$ & IS$\uparrow$ \\
  \midrule
  \textbf{DiT-XL/2} & 524.7 & 16.2 & 20.94 & 66.30 \\
  \midrule
  \textbf{DiC-XL} & 464.3 (228.7) & 84.2 & 15.32 & 93.55 \\
  \textbf{DiC-H} & 817.2 (388.4) & 53.3 & \textbf{12.89} & \textbf{101.78} \\
  \bottomrule
  \end{tabular}
  \vspace{-5pt}
  \caption{\textbf{Comparing DiCs against DiTs on ImageNet 512$\times$512 generation.} We compare models trained for 400K iterations with the standard setting of DiT. Please refer to the appendix for results from longer training iterations.}
  \label{tab:imagenet512}
  \vspace{-5pt}
\end{table}

\subsection{Scaling to Excellence}

In this section, we look at the scaled-up model (DiC-H) with increased training iterations, and the combination with diffusion techniques to discover the potential of \my.

\begin{table}[htbp] 
  \centering
  \setlength{\belowcaptionskip}{0cm}   
\begin{tabular}{lccc}
  \toprule
  \multicolumn{4}{l}{\bf{ImageNet} 256$\times$256, Scale Up, w/o cfg} \\
  \toprule
  Model & Training Steps & FID$\downarrow$ & IS$\uparrow$ \\
  \midrule
  \textbf{DiT-XL/2} & 2.4M & 10.67 & - \\
  \textbf{DiT-XL/2} & 7M & 9.62 & - \\
  \midrule

  \textbf{\my-H} & 400K & 11.36 & 106.52 \\
  \textbf{\my-H} & 600K & 9.73 & 118.57 \\
  \textbf{\my-H} & 800K & \textbf{8.96} & \textbf{124.33} \\
  \bottomrule
  \end{tabular}
  \vspace{-5pt}
  \caption{\textbf{Fast Convergence of \my.} The generation performance without classifier guidance of both models on ImageNet 256$\times$256 consistently improves as training progresses.}
  \label{tab:steps}
  \vspace{-10pt}
\end{table}

\noindent\textbf{Under DiT's plainest no-CFG setting.} Table \ref{tab:steps} presents the performance of \my-H on ImageNet 256$\times$256 as training progresses. The performance of \my-H improves steadily as the training progresses. At 600K steps, FID decreases to 9.73, matching the performance of DiT-XL/2 at 7M training steps; at 800K steps, \my-H shows the best performance with an FID of 8.96. This table demonstrates the fast convergence of \my-H, where both FID and IS consistently improve with training.

\begin{table}[!b]
  \centering
  \vspace{-15pt}
  \setlength{\belowcaptionskip}{0cm}   
  \begin{tabular}{lccc}
    \toprule
    \multicolumn{4}{l}{\bf{ImageNet} 256$\times$256, Scale Up, w/ cfg} \\
    \toprule
    Model & TP & BS$\times$Iter & FID$\downarrow$ \\
    \midrule
    \textbf{DiT-XL/2} & 66.8 & 256$\times$7M & 2.27 \\
    \textbf{U-ViT-H} & 63.9 & 1024$\times$500K & 2.29 \\
    \textbf{DiC-H (Ours)} & 160.8 & 256$\times$2M & \textbf{2.25} \\
    \bottomrule
  \end{tabular}
  \vspace{-5pt}
  \caption{\textbf{Generation performance of powerful baselines with classifier-free guidance.} TP stands for throughput. The proposed DiC is also performant on conditional generation.}
  \label{dicvsdit_scalecfg}
  \vspace{-15pt}
\end{table}

\begin{figure*}[htbp]
  \centering
  \includegraphics[width=\textwidth]{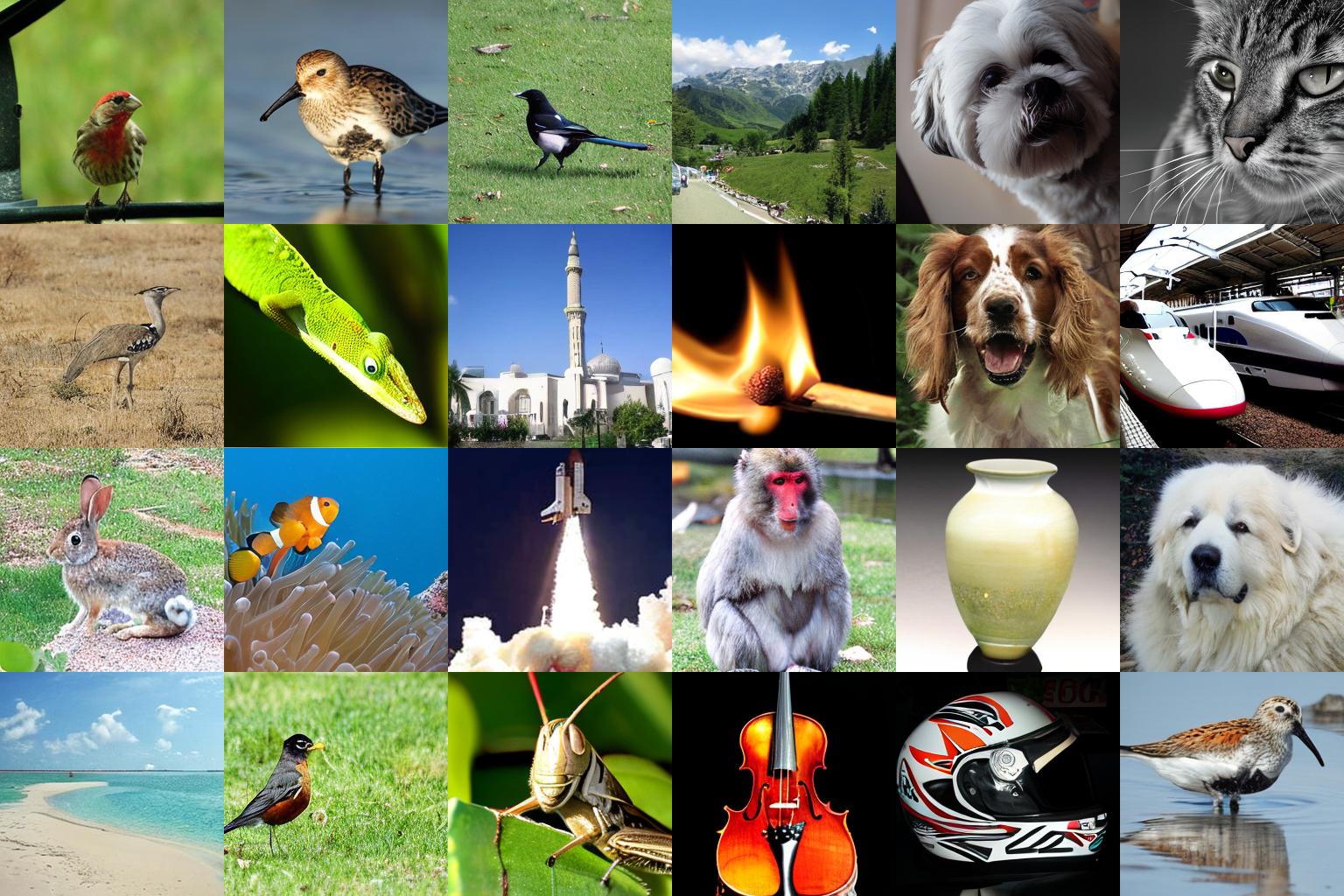}
  \vspace{-10pt}
  \caption{\textbf{Samples generated by DiC at 2M iterations.} The samples are generated following the setting of DiT, at $cfg=4$.}
  \label{fig:ipt}
  \vspace{-10pt}
\end{figure*}

\noindent\textbf{Towards excellence with classifier-free guidance.} Table \ref{dicvsdit_scalecfg} presents the comparison of various models' performance on ImageNet 256$\times$256 when training iteration is scaled up and classifier-free guidance is applied.

The baseline model DiT-XL/2 achieves a throughput of 66.8, processes a batch size of 256 for 7 million iterations, and yields an FID score of 2.27. The U-ViT-H model, while slightly lower in throughput at 63.9, uses a larger batch size of 1024, resulting in a slightly higher FID of 2.29. In comparison, our proposed DiC-H model shows a marked improvement when trained with a modest setting of batch size of 256 over 2 million iterations. DiC-H could achieve an FID of 2.25, with a throughput of 160.8. These results reveal that CNNs could reach performance that is no worse than Diffusion Transformers, and they have a great throughput advantage. Building upon this SiT flow-based framework, we also combine DiC with the State-of-the-Art technique of Representation Alignment (REPA)~\cite{repa,urepa} to achieve much faster convergence. The details are in the appendix.

\section{Conclusion}

In this work, we rethink the design of Conv3x3 in the context of diffusion models and introduced DiC, a purely 3x3-convolutional diffusion model. We initially explore various state-of-the-art diffusion architectures and find that the Encoder-Decoder U-Net structure perform the best. Building on this, we propose the sparse skip connection to reduce redundant block-wise connections and enhance efficiency of U-Net. In addition to architecture, we redesign the conditioning mechanism. We introduce stage-specific independent embeddings where each embedding is responsible for addressing conditioning within its assigned stage. We also propose mid-block condition injection and conditional gating, which are beneficial from empirical obvervations. Based on these improvements, our proposed Conv3x3 diffusion model not only achieves superior generative performance compared to Diffusion Transformers but also does so with significantly higher throughput. 

Above all, as an aspect that is always neglected in diffusion, we investigate the use of convolutions in diffusion models, and we hope this work could inspire further exploration into the use of convolutions in diffusion models.

\newpage

\noindent\textbf{Acknowledgement.} This work is supported by the National Key R\&D Program of China under grant No. 2022ZD0160300 and the National Natural Science Foundation of China under grant No. 62276007. We gratefully acknowledge the support of MindSpore, CANN, and Ascend AI Processor used for this research. 

{
    \small
    \bibliographystyle{ieeenat_fullname}
    \bibliography{main}
}

\clearpage
\setcounter{page}{1}
\maketitlesupplementary

\section{Additional Experiments}
\textbf{Further Details about Baselines.}  In Tab.~\ref{dicvsother}, most baselines including PixArt-$\alpha$, DiffiT~\cite{diffit}, and DiT-LLaMA~\cite{visionllama} are direct improvements over DiTs~\cite{dit}; U-ViTs~\cite{uvit} are published earlier to DiTs, and we also find some coincidence between the hyperparameters for model setup. The major difference between the architecture of U-ViTs and DiTs is the use of skip connections. DiTs totally eliminate the skips, maintaining a clear isotropic architecture. The results of DiT-LLaMA is replicated by us because it fails to report an official result for 400K iterations under the DiT setting. This omission is strange to us because DiTs report the 400K results for smaller models when compared with DiT.

\noindent\textbf{Credit.} Baseline performance statistics in Tab.~\ref{dicvsother} are from~\cite{udit}, a work that measures the capability of Diffusion Transformers under the aligned standard setting of DiT.

\noindent\textbf{DiT Combined with U-Net.} We also conducted the experiment that combines DiT transformer block with the U-Net architecture, shown in Tab.~\ref{tab:appendixudit}. In contrast, U-Nets could bring more improvements to field-limited ConvNets.

\begin{table}[htbp] 
  \centering
  \setlength{\belowcaptionskip}{0cm}   
\begin{tabular}{lccc}
  \toprule
  \multicolumn{4}{l}{\bf{ImageNet} 256$\times$256, 200K, cfg=1.5} \\
  \toprule
  Model & G FLOPs & FID$\downarrow$ & IS$\uparrow$ \\
  \midrule
  \textbf{DiT-XL/2} & 118.6 & 12.96 & 94.26 \\
  \textbf{DiT+U-Net} & 117.5 & \textbf{11.03} & \textbf{104.92} \\
  \bottomrule
  \end{tabular}
  \vspace{-5pt}
  \caption{\textbf{Improvements of U-Net on DiT.} The improvements of U-Net on transformers are not as large as on ConvNets.}
  \label{tab:appendixudit}
  \vspace{-5pt}
\end{table}

\noindent\textbf{More Traing Iterations on ImageNet 512$\times$512.} We have extended the training iterations for some limited number of iterations, shown in Tab.~\ref{tab:appendix512long}. Both DiC-XL and DiC-H could outperform DiT-XL/2 at much fewer training iterations while maintaining a speed advantage. As is shown in Tab.~\ref{tab:appendix512_3m}, DiC models could also perform better than DiT-XL at 3M iterations, besides the complexity and speed advantage of Conv3x3 especially on larger images.

\begin{table}[htbp] 
    \centering
    \setlength{\belowcaptionskip}{0cm}   
  \begin{tabular}{lccc}
    \toprule
    \multicolumn{4}{l}{\bf{ImageNet} 512$\times$512, Scale Up, w/o cfg} \\
    \toprule
    Model & Training Steps & FID$\downarrow$ & IS$\uparrow$ \\
    \midrule
    \textbf{DiT-XL/2} & 1.3M & 13.78 & - \\

    \midrule
  
    \textbf{\my-XL} & 600K & 13.64 & \textbf{102.63} \\
    \textbf{\my-H} & 400K & \textbf{12.89} & 101.78 \\
    \bottomrule
    \end{tabular}
    \vspace{-5pt}
    \caption{\textbf{Fast Convergence of \my.} DiC models could achieve better performance at much fewer training iterations on ImageNet 512x512.}
    \label{tab:appendix512long}
    \vspace{-5pt}
  \end{table}

\begin{table}[htbp] 
    \centering
    \setlength{\belowcaptionskip}{0cm}   
    \begin{tabular}{lcccc}
      \toprule
      \multicolumn{5}{l}{\bf{ImageNet} 512$\times$512, 3M, cfg=1.5} \\
      \toprule
      Model & G FLOPs (Wino.) & TP & FID$\downarrow$ & IS$\uparrow$ \\
      \midrule
      \textbf{DiT-XL/2} & 524.7 & 16.2 & 3.04 & 240.82 \\
      \midrule
      \textbf{DiC-XL} & 464.3 (228.7) & 84.2 & 3.04 & 271.77 \\
      \textbf{DiC-H} & 817.2 (388.4) & 53.3 & \textbf{2.96} & \textbf{293.54} \\
      \bottomrule
      \end{tabular}
    \vspace{-5pt}
    \caption{\textbf{The performance of \my~on larger images at 3M iterations.} DiC models could achieve better performance and higher throughput than DiT-XL on ImageNet 512$\times$512.}
    \label{tab:appendix512_3m}
    \vspace{-5pt}
  \end{table}

\begin{figure}[htbp]
    \centering
    \includegraphics[width=0.45\textwidth]{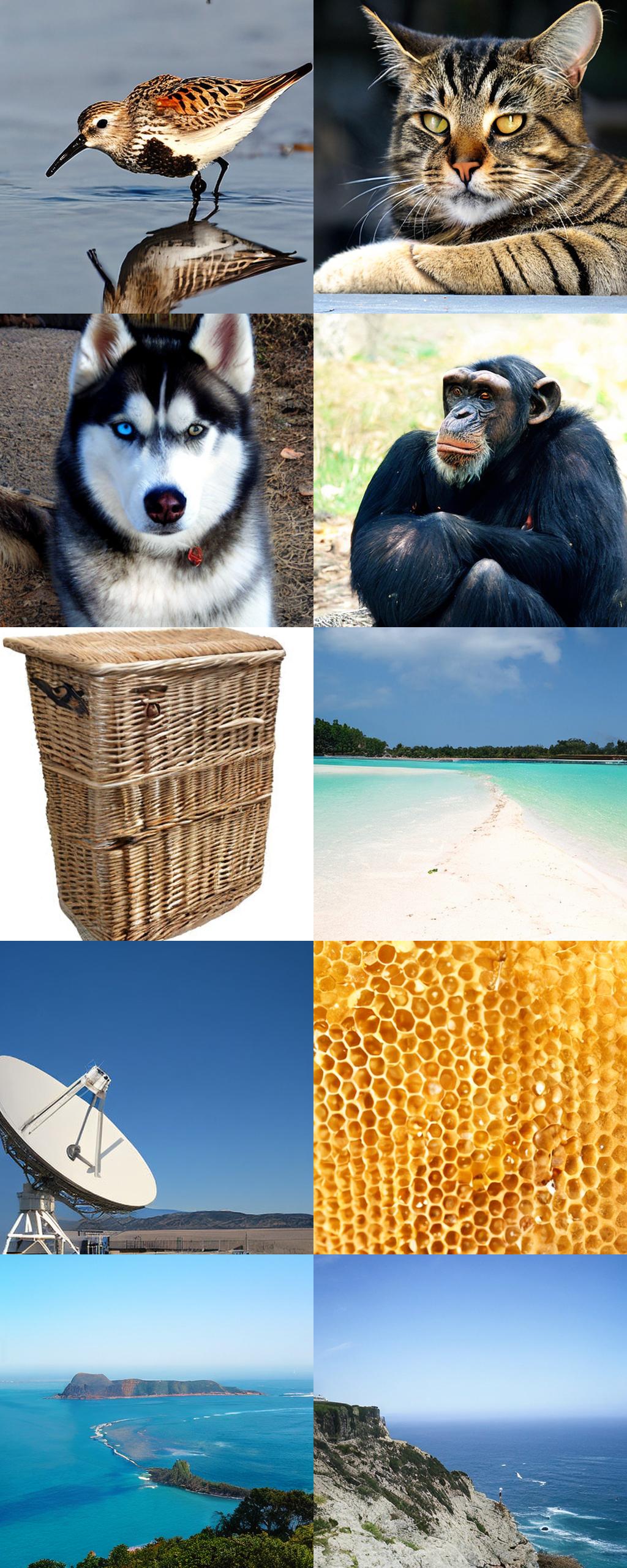}
    \vspace{-5pt}
    \caption{\textbf{512x512 Samples generated by DiC at 1M iterations.} The samples are generated following the setting of DiT, at $cfg=4$. Best viewed on screen.}
    \label{fig:ipt512}
    \vspace{-10pt}
\end{figure}

\noindent\textbf{Visual Results from ImageNet 512x512.} In Fig.~\ref{fig:ipt512}, we also present the samples generated by DiC-XL, trained for 1M iterations. The samples are generated with the setting of DiT.

\begin{figure}[!t]
  \centering
  \includegraphics[width=0.4\textwidth]{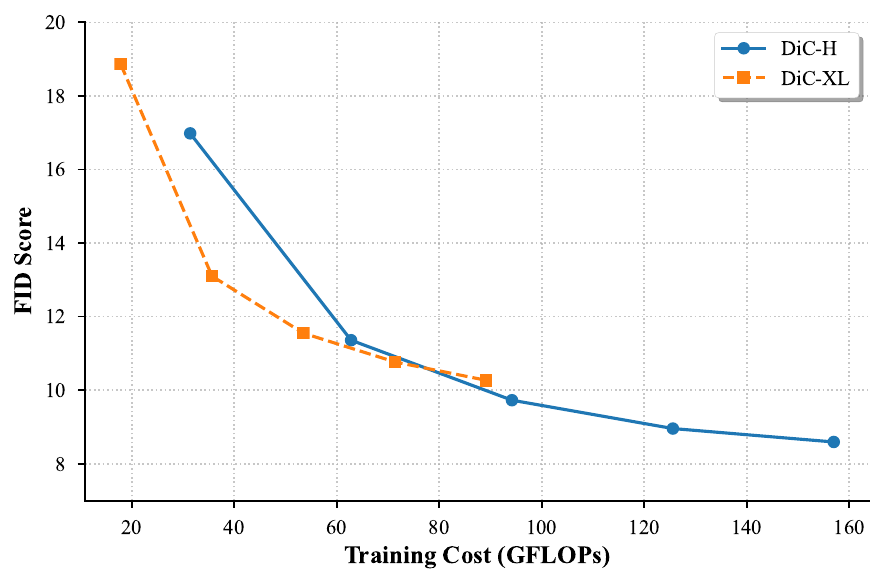}
  \vspace{-10pt}
  \caption{\textbf{The scaling curve of DiC training.} The FID scores are recorded once 
every 200K iterations in the first 1M training iterations. The scaling effect of DiC as model gets larger is obvious from the plot.}
  \label{fig:scaling}
  \vspace{-10pt}
\end{figure}

\noindent\textbf{Scaling plots.} We visualize the scaling curve of DiC-XL and DiC-H as shown in Fig.~\ref{fig:scaling}.

\noindent\textbf{Comparing with more architectures.} Apart from the architectures mentioned in Tab.~\ref{dicvsdit}, we are aware of some other competitive architectures including Simple Diffusion~\cite{simplediffusion}, RIN~\cite{rin}, EDM2~\cite{edm2}, and HDiT~\cite{hourglass}. However, we find difficulty in comparing these methods with DiC: DiC is mainly focused on 256-sized latent diffusion, following DiT~\cite{dit} and SiT~\cite{sit}; these work, on the other hand, focuses on large (mostly pixel-space) diffusion; and they require large training costs to reach SOTA FIDs (e.g. EDM2 requires the training cost of 939.5-2147.5M img, which is around 4M to 8M iterations in our setting). We try to align HDiT, RIN, and EDM2 to our setting (under the training framework of DiT; in order to keep FLOPs aligned with DiC for fair comparison, we increase the depths and widths of these models). Results turns out that these methods either converges slowly (RIN, EDM2) or completely fails (HDiT).

\noindent\textbf{Details regarding Representation Alignment on DiC.}  We consider applying Representation Alignment (REPA)~\cite{repa} (using its variant U-REPA~\cite{urepa} tailored for U-Net) to achieve faster convergence. We use the standard training hyperparameter for REPA. For sampling, we use $cfg=1.8$ for SDE, and $cfg=2$ for ODE, both equipped with guidance interval (following the default setting of the official codebase). Amazingly, DiC-XL could reach an FID of 1.74 after 1M training steps with the help of REPA, as shown in Tab.~\ref{tab:appendix_repa}.

\begin{table}[htbp] 
    \centering
    \setlength{\belowcaptionskip}{0cm}   
  \begin{tabular}{lccc}
    \toprule
    \multicolumn{4}{l}{\bf{ImageNet} 256$\times$256, REPA} \\
    \toprule
    Model & Training iter & Sampling & FID$\downarrow$ \\
    \midrule
    \textbf{\my-XL+U-REPA} & 1M & ODE & 1.74 \\
    \textbf{\my-XL+U-REPA} & 1M & SDE & 1.75 \\
    \bottomrule
    \end{tabular}
    \vspace{-5pt}
    \caption{\textbf{Fast Convergence of DiC with the help of REPA.} DiC-XL could achieve 1.75 FID after 1M training iterations.}
    \label{tab:appendix_repa}
    \vspace{-5pt}
  \end{table}

\end{document}